\documentclass{article}

\usepackage[]{graphicx}
\usepackage[]{color}
\usepackage{geometry}
\usepackage{natbib}

\newenvironment{knitrout}{}{} % an empty environment to be redefined in TeX

\makeatletter
\def\maxwidth{ %
  \ifdim\Gin@nat@width>\linewidth
    \linewidth
  \else
    \Gin@nat@width
  \fi
}
\makeatother

\definecolor{fgcolor}{rgb}{0.345, 0.345, 0.345}

\usepackage{framed}
\makeatletter
 {\par\unskip\endMakeFramed%
 \at@end@of@kframe}
\makeatother

\title{Effects of Sampling Methods on Prediction Quality. The Case of Classifying Land Cover Using Decision Trees.}
\author{Ronald Hochreiter \and Christoph Waldhauser}
\date{May 2014}

\begin{document}

\maketitle

\begin{abstract}
Clever sampling methods can be used to improve the handling of big data and increase its usefulness. The subject of this study is remote sensing, specifically airborne laser scanning point clouds representing different classes of ground cover. The aim is to derive a supervised learning model for the classification using CARTs. In order to measure the effect of different sampling methods on the classification accuracy, various experiments with varying types of sampling methods, sample sizes, and accuracy metrics have been designed. Numerical results for a subset of a large surveying project covering the lower Rhine area in Germany are shown. General conclusions regarding sampling design are drawn and presented.
\end{abstract}

%\COMPSTATKeywords{Machine Learning, Sampling, Decision Tree, Airborne Laser Scanning}

\section{Introduction}

% The advent of big data has revived a criticism statistics as a discipline has always faced: uncertainty introduced by sampling.
% While in many applications of statistics, total measurments are prevented by sheer volume and expense, in a big data scenario the situation is different.
% There, a lot of effort was put into generating large amounts of data in the first place.
% Statistician's suggestions to not use all of that data and rather draw a sample out of it for further analysis often enough provokes an outcry by the data provider.

% Motivation: can't classify everything by hand, need a clever selection of what to classify. To stratify or not to stratify? Weights?

In this paper we seek to understand how clever sampling can be used to improve the handling of big data and increase its usefulness.
Our case comes from remote sensing and contains airborne laser scanning point clouds representing different classes of ground cover.
The aim is to derive a supervised learning model for the classification using CARTs.
This paper is organized as follows. We will first briefly review the state of the art for the fields of airborne laser scanning, classification trees and stratified sampling.
We will then describe our data set and the experimental setup.
In Section \ref{ronald.hochreiter:sec:results} we present our results. After discussing them we conclude with suggestions for further research.

\section{Classification of Airborne Laser Scan Point Clouds}

The character of land surveying has changed dramatically in recent decades.
The availability of powerful lasers scanning, surveying the ground from airborne platforms provides very accurate and timely measurements of ground cover.
These airborne laser scans (ALS) produce geo-coded point clouds of laser return echos with resolutions in a decimeter scale.
However, the resulting massive amounts of data -- surveying planes can easily cover vast stretches of land -- require specialized handling.
We are now going to discuss the specifics of ALS.
Then we review four sampling procedures in the context of ALS.
Finally we explain the usage of the popular CART algorithm for automatically deriving classifications of ground cover.

\subsection{Airborne Laser Scanning}

% why use ALS (applications)
% what does the data structure look like (i.e. features, manual classification)

Before the advent of ALS, land surveying was very cumbersome and involved traveling target areas, setting up measurement points.
The situation improved somewhat due to the increased usage of aerial photography.
These photos, however, are often ambiguous and it is difficult to tell three-dimensional structures from them, even when using stereo photography.

Airborne laser scanning, on the other hand offers a number of advantages.
An airplane or helicopter flies over a designated area usually stripwise.
A downward looking laser array emits beams and records the echos.
Modern arrays use rotating mirrors to deflect the beam and scan an area perpendicular to the flight path.
Also, these arrays are capable of recording the entire spectral characteristics of multiple echos.

From these raw return signal characteristics and the GPS coordinates of the plane, every signal is translated into points on (or above) the ground.
Further point features can be computed from the signal using various neighborhood averaging techniques.
For an overview and a systematic organization of available point features, see \citep{otepka2013georeferenced}.
From this follows that every point is described by a high-dimensional feature vector.

These point clouds can be used instantly to compute digital terrain and surface models.
In order to use them for other purposes, e.g. the identification of ground cover, classification models are needed.
We are going to describe them in the next subsection.

\subsection{Classifying ALS point clouds using CARTs}

% why manual classification should be replaced
% how carts work
% how they are used in classification of ground cover (i.e. how we use them)

The deriving of types of ground cover from either aerial photos or ALS point clouds, is essentially a quite simple but time-consuming task.
It involves investigating the aerial photos and use ones experience to classify a given area.
For instance, looking at a certain shape in the point cloud and cross-referencing it with data from the photo, the human classifier can classify the related points to be coniferous forest or a road.
Processing hundreds of acres in this manner is time consuming, expensive and error prone.
It is therefore the aim of ongoing research to develop automatic classifiers.
There has been some success in that matter. 
See e.g. \citep{waldhauser14} for an overview of current research.
A summary of the state of ALS classification needs to point to still rather high misclassification rates and difficulties in classifying certain types of ground cover.

Automated classification tasks, in general, can be supervised or unsupervised.
The latter works entirely autonomously and draws all information from the available data.
The former requires manually created training data to learn its model from.
An example of a supervised learning algorithm are classification and regression trees (CARTs) \citep{breiman1984classification}.
And it is CARTs that have been used with great success in the past \citep{hothorn2006unbiased,pal2003assessment,waldhauser14}.
There, a large number of manually classified points are used to train and evaluate the classification of ground cover.
For evaluation purposes, the data set is artificially split into training and test data.
In practical applications, however, the training area needs to be determined using statistical sampling techniques.
In the next subsection we are going to review four different sampling procedures that we find useful in this context.

\subsection{Sampling in the ALS context}

% stehman text on sampling in geo contexts
% how sampling looks practically (i.e. sparse classes)

The most natural way of obtaining a slice of a data set is to simply take the first, say 10,000 cases.
This, however, is not a random sample and it is rather unlikely that it will be representative enough to train a classifier from it.
A better approach and perhaps the second most natural thing to do is a simple random sample.
Since the extent of the entire data set is known when sampling, every point has the same chance of being included in the sample.
The key advantage of this approach is obtaining a sample that is representative of the data set.

However, there is also a big disadvantage: rare classes that are only found sparingly throughout the data set might not be sampled at all.
Then of course the classifier has no chance of learning the characteristics of these classes and will not be able to assign any data into these classes during classification.
For instance, consider that a moderately small data set contains 2 million points and a single tree might be made up of as little as 500 points.
For being able to classify this particular kind of tree, enough of its points need to end up in the sample.
For simple random samples this means that large samples are required to ensure sampling of rare classes with reasonable probabilities.

Another approach originally invented for surveys among humans is stratified sampling \citep{anganuzzi1993post,holt1979post,little1993post}.
There, a small simple random sample is taken from each class, no matter how rare it is.
Obviously, the resulting sample is not representative of the population anymore.
But presence of points from rare classes is guaranteed.
In the ALS context, stratified sampling has been used numerous times e.g.\ \citep{joy2003non,stehman1996estimating}.

When using samples that are not representative of the population they were drawn from, care must be taken.
The canonical method is to compute sampling weights that correct for the altered composition of the sample.
As CARTs actively use the distribution of classes when estimating class probabilities, the sampling weights are used to compute correct priors.
Alternatively, CARTs' computation of priors can be circumvented and the true priors be specified. 

In order to examine the effect of these sampling plans, we used a data set from a real surveying project in a series of experiments.
Both are being described in the next section.

\section{Data \& Method}

% Descriptives of data set (1 table)
% Description of experimental setup (table/graphic?)
% Description of quality assessment metrics (TO-MCR, CB-MCR, Kappa)

In this section we are going to introduce the data set we used in our analysis as well as the experimental setup.
Our data set is a small subset of a large surveying project covering the lower Rhine area in Germany.
Our subset was selected for its representativeness of the region.\footnote{All computations were done in R \citep{r-core} using \emph{lasr} \citep{lasr} for point cloud processing,  \emph{rpart} \citep{rpart} for CARTs and \citep{ggplot2} for visualization.}
It contains 2,872,488 points.
The data set was manually annotated using photo interpretation and ALS point clouds.
Table \ref{ronald.hochreiter:tab:desc} contains the distribution of points in the classes that were observed.

% latex table generated in R 3.1.0 by xtable 1.7-3 package
% Tue May 13 20:24:26 2014
\begin{table}[ht]
\centering
\begin{tabular}{lr}
  \hline
Class of ground cover & Frequency \\ 
  \hline
undefined & 959 \\ 
  ground & 2401914 \\ 
  gravel & 1903 \\ 
  asphalt & 20301 \\ 
  decideous forest & 175103 \\ 
  building roofs & 1383 \\ 
  walls/buildings &  13 \\ 
  water & 61362 \\ 
  cars and other moving objects & 3912 \\ 
  temporary objects & 1411 \\ 
  bridges & 173774 \\ 
  power poles & 231 \\ 
  bridge cables & 16240 \\ 
  road protection fence & 11169 \\ 
  bridges construction & 1819 \\ 
  cement/concrete & 936 \\ 
  error class &  58 \\ 
   \hline
\end{tabular}
\caption{Distribution of ground cover classes in the data set} 
\label{ronald.hochreiter:tab:desc}
\end{table}

As can be seen from Table \ref{ronald.hochreiter:tab:desc}, the distribution is highly skewed with roughly 85\% being ground.
There are also some very rare classes that will be hard to sample and learn, e.g.\ walls.

In order to measure the effect of sampling methods on the classification accuracy we designed experiments with varying types of sampling methods, sample sizes, and accuracy metrics.
The two types of sampling methods were simple random samples (without replacement) and stratified samples.
While the first one is straight forward, the second type of sampling method deserves explanation.
The sampling algorithm tried to sample $s$ points from each class.
If a class had less than twice as many points, it would only sample half of those points.

This sampling method then also dictated sample sizes.
In total 8 different sample sizes were used.
Table \ref{ronald.hochreiter:tab:ss} gives an overview of the sample sizes used in the experiments.

% latex table generated in R 3.1.0 by xtable 1.7-3 package
% Tue May 13 20:24:26 2014
\begin{table}[ht]
\centering
\begin{tabular}{lr}
  \hline
Sample & Points \\ 
  \hline
S1 & 84288 \\ 
  S2 & 72524 \\ 
  S3 & 57783 \\ 
  S4 & 39251 \\ 
  S5 & 18472 \\ 
  S6 & 4268 \\ 
  S7 & 906 \\ 
  S8 & 471 \\ 
   \hline
\end{tabular}
\caption{Sample sizes used in experiments. The entire data set contains 2.9 million points.} 
\label{ronald.hochreiter:tab:ss}
\end{table}

In order to gain an estimate of the variability of our measurements, we drew fifty samples of each size using each type of sampling method.

These samples were then used to train a classifier using the CART framework introduced above.
The so obtained classifier was then used to classify the remainder of the data set.
When learning off stratified samples, the CART algorithm was given (a) no further information on class priors, (b) post-stratification case weights\footnote{Post-stratification was computed using \emph{survey} \citep{lumley2004analysis}.}, and (c) the true class distributions.
We therefore have four (simple random samples and stratified samples in the three fashions outlined above) different sampling methods to compare.

To gauge the quality of a classification we used three different metrics all based on cross classification error matrices $M$.
The simplest metric is the total misclassification rate ($MCR_{Total}$) that is defined as 

\[
MCR_{Total} = 1-\frac{\sum \mathrm{diag}(M)}{N}
\]

with $N$ being the total number of points.
A distinct property of the total MCR is that the misclassification of rare classes has only a small influence.
Whether this is considered a bug or a feature depends on the context.
In bids to accurately classify even rare classes, the total MCR can be misleading.

A approach that takes rare classes into account is the class-based MCR ($MCR_{Class}$).
There, the misclassification is not averaged over the entire data set but per class:

\[
MCR_{Class} = \sum 1-P(a)
\]

with $P(a)$ being the proportion of correctly classified points in that class.
Another quality metric is Kohen's Kappa ($\kappa$) \citep{stehman1996estimating}.
It is given as 

\[
\kappa = \frac{P(a) - P(e)}{1-P(e)}
\]

with $P(e)$ is the product of marginal proportions for that class.
We computed each of these metrics for every sample size and sampling method combination.
The results of these computations are given in the next section.

\section{\label{ronald.hochreiter:sec:results}Results}

% Boxplots comparing classification stability
% Line charts comparing usability of small samples
% Boxplots (linecharts?) comparing model training time for sample sizes

In order to assess the effects of sampling methods on the classification accuracy, we conducted a series of experiments.
In the following we will present the results from our computations.
Figure \ref{mcr} displays the relationships between sample size and classification accuracy for the four sampling methods.
We depicted only the three smallest sample sizes as they---obviously---exhibit the largest differences, but larger sample sizes confirm the overall trend.

For the overall misclassification rate, simple random samples outperform all other classification approaches clearly.
It is also notable, that sample size has no significant degrading effect on simple random samples using this metric.

Class-based misclassification rates reverse this picture. Using that metric, simple random samples perform very poorly and stratified samples provide better, consistent results.
Among the stratified samples, interestingly, the method of not providing the CART algorithm with any further information on the priors or sample composition outperforms the other methods.

Finally, turning to $\kappa$, simple random samples once again deliver the best performance by achieving the greatest agreement.
Among the stratification methods, post-stratification and prior specification both perform slightly better than their uninformed brother.

In this section, we presented the results from our experiments.
They were 50 times bootstrapped each.
It is remarkable to see that, while simple random sampling beats any other method by far in two of three metrics, this is not true when using the class-based misclassification metric. There, working with stratified samples not correcting for the stratification delivers best performance. 
In the next section we are going to discuss these findings in greater detail.

\begin{knitrout}
\definecolor{shadecolor}{rgb}{0.969, 0.969, 0.969}\color{fgcolor}\begin{figure}[]

{\centering \includegraphics[width=\maxwidth]{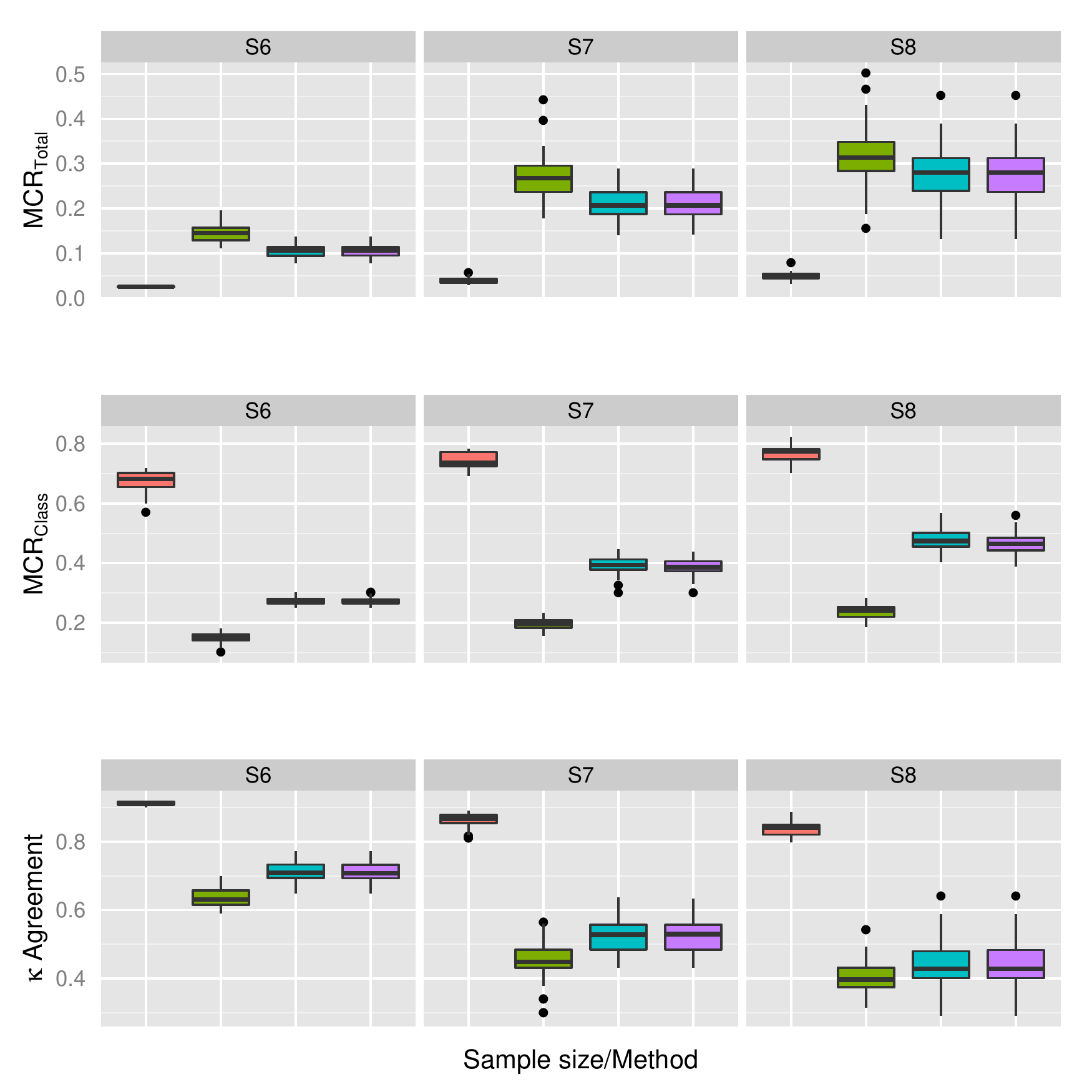} 

}

\caption[\label{mcr}Quality of the obtained prediction for different sample sizes and sampling methods measured in three metrics]{\label{mcr}Quality of the obtained prediction for different sample sizes and sampling methods measured in three metrics. Results are 50 times bootstrapped each. Sampling methods per panel, from left to right: simple random sample, stratified sample, stratified with post-stratification weights, stratified with priors specified.\label{fig:to-mcr}}
\end{figure}

\end{knitrout}

\section{Discussion \& Conclusion}

We have compared the prediction quality using CARTs on land surveying laser scan data, given different sample sizes and sampling methods.
Prediction quality was measured using three different metrics putting varying degrees of emphasis on overall correctness or correctness per class.
Bootstrapping the results fifty times, we find that simple random samples provide the best overall classification quality.
When using stratified samples in that context, it is important to specify either post-stratification weights or true class probabilities to achieve better results.

However, when defining classification quality as average per class, we find that stratified samples achieve much lower misclassification rates.
Interestingly, this is even the case when the CART algorithm is not provided with true or assumed class probabilities.
More precisely, specifying class probabilities or post-stratification weights to undo the stratification process lead to worse results.
This is unexpected, as correcting the altered sample composition introduced by stratification is thought to be canonical \citep{doss1979exact} and has been shown to improve classification even with land surveying data above.

As CART-based analysis is becoming ever more widespread, also in survey contexts, our findings warrant caution when integrating the sampling design with the CART analysis.

\bibliographystyle{plainnat}
\bibliography{hw-lit.bib}

\end{document}